%% file: main.tex
\documentclass{svproc}
\usepackage{url}
\usepackage{graphics} 
\usepackage{epsfig}
\usepackage{amsmath}
\usepackage{subcaption}

\usepackage{amssymb}
\usepackage{dsfont}

\DeclareMathOperator*{\argmax}{arg\,max}

\begin{document}
\mainmatter              
\title{Online vs. Offline Adaptive Domain Randomization Benchmark}
%
\titlerunning{Online vs. Offline Adaptive Domain Randomization Benchmark}  
%
\author{
Gabriele Tiboni\inst{1} \and
Karol Arndt\inst{2} \and
Giuseppe Averta\inst{1} \and \\
Ville Kyrki\inst{2} \and
Tatiana Tommasi\inst{1}
}
\authorrunning{Tiboni et al.} 
%
\tocauthor{Gabriele Tiboni, Karol Arndt, Giuseppe Averta, Ville Kyrki, and Tatiana Tommasi}
%
\institute{Politecnico di Torino, Torino 10129, Italy,\\
\email{first.last@polito.it}
\and
Aalto University,
Espoo 02150, Finland,\\
\email{first.last@aalto.fi}}

\maketitle              

\begin{abstract}
Physics simulators have shown great promise for conveniently learning reinforcement learning policies in safe, unconstrained environments. However, transferring the acquired knowledge to the real world can be challenging due to the reality gap. To this end, several methods have been recently proposed to automatically tune simulator parameters with posterior distributions given real data, for use with domain randomization at training time. These approaches have been shown to work for various robotic tasks under different settings and assumptions.
Nevertheless, existing literature lacks a thorough comparison of existing adaptive domain randomization methods with respect to transfer performance and real-data efficiency.
In this work, we present an open benchmark for both offline and online methods (SimOpt, BayRn, DROID, DROPO), to shed light on which are most suitable for each setting and task at hand.
We found that online methods are limited by the quality of the currently learned policy for the next iteration, while offline methods may sometimes fail when replaying trajectories in simulation with open-loop commands.
The code used will be released at~\url{https://github.com/gabrieletiboni/adr-benchmark}.
%
\end{abstract}

\section{Introduction}
\label{sec:introduction}
\input{sections/01_introduction.tex}

\section{Related works}
\label{sec:related_works}
\input{sections/02_related_works.tex}

\section{Methodology}
\label{sec:methodology}
\input{sections/03_methodology.tex}

\section{Results}
\label{sec:results}
\input{sections/04_results.tex}

\section{Conclusions}
\label{sec:conclusions}
\input{sections/05_conclusions.tex}

\section{Acknowledgments}
We acknowledge the computational resources generously provided by HPC@POLITO and by the Aalto Science-IT project.

\bibliographystyle{spmpsci}
\bibliography{biblio.bib}



\end{document}

%% file: sections/01_introduction.tex
Recent advancements in the field of deep Reinforcement Learning (RL) have shown promising results for many robotic applications, by allowing to solve tasks through simple trial-and-error. On one side, this has opened a new path for collaborative robots that autonomously learn complex skills. On the other, learning through interactions with the environment still requires a substantially large number of training data and poses important safety risks~\cite{kober2013reinforcement}.
To this end, researchers have adopted strategies to conveniently train robots for safe optimal control, such as learning from human demonstrations or by interacting with unconstrained simulated environments.
There has been a growing interest in the latter approach in recent years, driven by several success stories of robotic tasks being learned exclusively in simulation, from robotic in-hand manipulation~\cite{ADRrubikscube}, to drone flight~\cite{LearningFromRandomizedSimulators} and locomotion for quadruped robots~\cite{locomotion}.
%
Albeit promising, however, the effectiveness of a sim-to-real transfer strongly depends on the quality and the accuracy of the physics engines that simulate the underlying scenario.
Indeed, directly transferring robot policies learned in a poorly calibrated simulator---a.k.a. the \textit{source domain}---usually leads to low performance on real hardware~\cite{AusersGuideForCalibratingSimulators}---the \textit{target domain}. Such discrepancy is commonly known as the \textit{reality gap} and represents the main limitation for RL in robotics. Generally speaking, the reality gap may be attributed to a combination of factors, such as mismatched dynamics models and unmodeled phenomena. Additionally, humidity and temperature changes, together with wear and tear of robotic setups, often lead to dynamically changing real-world environments.

To bridge the reality gap and learn more generalizable RL agents, several works have proposed to train robots over varying simulated environments, i.e. by randomizing dynamics parameters according to a predefined probability distribution. This technique, known as \textit{Domain Randomization} (DR), has demonstrated to be a promising method to learn robust robot policies that can be directly transferred to the target domain~\cite{zhao2020sim}. More specifically, DR requires designing a source domain distribution over physical parameters that allow for both training of a single well-performing policy on varying dynamics, and for efficient transfer to the real system. This problem attracted researchers to move from point-estimate tuning techniques to full posterior distribution inference for use with Domain Randomization. A recent survey~\cite{LearningFromRandomizedSimulators} refers to such methods as \textit{Adaptive Domain Randomization} (ADR), as the primary goal is to automatically infer source domain distributions, as opposed to using manually engineered static distributions.
\begin{figure*}[t!]
\includegraphics[width=\textwidth]{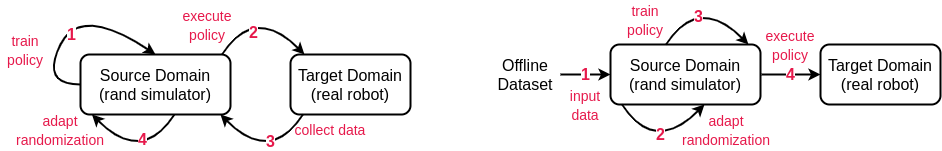}
    \caption{Conceptual illustration of online (left) and offline (right) adaptive domain randomization.}
    \label{fig:on-vs-off_scheme}
    \vspace{-0.5cm}
\end{figure*}
%
Despite the plethora of recently published ADR methods \cite{ADRrubikscube,BayesSim,simopt2019,droid2021,bayrn2021,dropo2022}, their problem setting, robotic tasks, and imposed assumptions often differ. 
Furthermore, each method presents superior performance w.r.t. their proposed baselines, but these have been hardly ever compared to each other. In accordance with~\cite{LearningFromRandomizedSimulators}, we claim that while these works have critically different assumptions, the existing literature lacks a thorough comparison of real-world transfer performance of existing ADR methods. A similar comparison may shed light on which methods are more suitable for certain problem settings than others, besides investigating current limitations for future studies.

With this purpose, our work presents an open benchmark of four ADR methods (SimOpt\cite{simopt2019}, BayRn \cite{bayrn2021}, DROID \cite{droid2021}, DROPO \cite{dropo2022}) tasked with sim-to-sim transfers under noisy conditions and unmodeled phenomena. We focus on target domain performance as the metric of interest, rather than inspecting the accuracy of the inferred parameter posteriors. To further investigate each method's limitations, we explore how performances vary according to the amount of data collected from the target domain, which is often critically limited and expensive to obtain in robotics scenarios. In particular, we distinguish between \textit{online} approaches~\cite{simopt2019,bayrn2021}, which iteratively gather on-policy data from the target domain, and \textit{offline} approaches~\cite{droid2021,dropo2022}, which rely on fixed offline datasets (see Figure \ref{fig:on-vs-off_scheme}). For example, offline methods can be fed with human demonstrations which feel more natural to non-expert users when interacting with the real setup, as opposed to rolling out RL policies iteratively with careful supervision. However, online approaches might reach higher performance in the long run by adapting to the target domain through multiple iterations.
To the best of our knowledge, this is the first work aiming to provide insights into how algorithms from the two aforementioned ADR settings perform along a variety of axes and tasks.

The contributions of this work are: (1) shedding light on the critically different assumptions and limitations of online vs. offline ADR methods; (2) providing a thorough performance analysis over four ADR methods w.r.t. target domain performance and data efficiency; (3) investigating how different data collection strategies affect the performances of offline methods.

%% file: sections/02_related_works.tex
Domain Randomization (DR) has been gaining wider adoption in recent years as a method for sim-to-real transfer, driven by its ease of implementation and effectiveness.
Antonova et al.~\cite{pivoting} were among the first to show how robust RL policies may be learned by randomizing parameters in the model of the training environment.
Soon after, several works have demonstrated impressive robotic skills trained exclusively in simulation with DR, such as 
solving the Rubik's cube~\cite{ADRrubikscube}, learning locomotion tasks~\cite{locomotion}, autonomously opening a door~\cite{tactilesensory}, or pushing objects with arbitrary friction properties~\cite{dynamics2018}. Valassakis et al.~\cite{Valassakis} have further investigated how DR compares to other sim-to-real transfer techniques for simple robotic manipulation tasks. Moreover, a recent study~\cite{chen2022understanding} has attempted to provide a theoretical framework around DR for dynamics to explain its impressive empirical results.
While not explored in this work, the same concept has also been applied in the field of deep RL for vision-based policies, by randomizing appearance properties of the simulated environment~\cite{DRvision2017iros,DRvision2017rss,James2017}.

As the complexity of the task increases, 
the main challenge lies in how to design source domain distributions to both encompass real-world physical parameters and make up for mismatched dynamics models~\cite{howtopickdrdistributions}.
When fixed uniform distributions are picked, the approach is generally referred to as Uniform Domain Randomization (UDR). UDR may be the result of prior task-specific knowledge or tedious back-and-forth tuning until the desired target performance is achieved.

To relax the manual engineering efforts associated with this approach, Adaptive Domain Randomization (ADR) methods attempt to automatically infer the source domain distribution.
Early attempts by Rajeswaran et al.~\cite{epopt} in this direction demonstrated that physical parameters could be inferred 
by collecting data from the target domain, despite showing experiments in simulation only.
More recent works gained popularity by applying a similar concept to actual sim-to-real transfer tasks, and using the inferred source domain distribution for domain randomization at training time: SimOpt~\cite{simopt2019} infers dynamics parameters by minimizing a discrepancy metric between real and simulated trajectories, while BayRn~\cite{bayrn2021} directly optimizes for real-world returns by framing inference as an end-to-end Bayesian Optimization problem. While training always happens in simulation, these methods require iterative target domain data collection---as they roll out the currently trained policy to ask for real-world feedback---potentially posing data efficiency concerns and assuming the real hardware to be available at training time. We refer to these as \textit{online} ADR methods.

\vspace{-6pt}
An opposite line of work has been recently proposed by exploiting offline fixed datasets to infer source domain distributions: DROID~\cite{droid2021} minimizes joint torques when replaying real-world trajectories in simulations, while DROPO~\cite{dropo2022} infers dynamics parameters given real-world transitions in a maximum-likelihood framework. These methods make no assumptions on how the given dataset has been collected, making them suitable for use with human demonstrations---which are often conveniently available in many robotic setups---or data collected by previously trained policies for other tasks. By nature, offline ADR approaches thrive on safety-critical tasks where rolling out multiple intermediate policies may be too expensive. On the other hand, these may fall short to readily adapt to the target domain given limited data, as opposed to online methods.

\vspace{-6pt}
Our work aims to provide the first empirical insights to support or re-evaluate the above premises, by comparing the four previously described methods under noisy conditions and unmodeled phenomena in simulation, by means of target domain performance and data efficiency. We chose these methods as they all leverage feed-forward network architectures, gradient-free optimization for non-differentiable simulators, and parametric source domain distributions.
We acknowledge some similarities with~\cite{AusersGuideForCalibratingSimulators}, a recent open benchmark on system identification for physics simulator tuning. However, this study purely focused on point-estimate parameter inference---i.e. system identification---and did not discuss the data efficiency requirements. In addition, each algorithm is tested in direct parameter estimation under noiseless and perfectly modeled conditions.

\vspace{-6pt}
As the lack of a common benchmark for DR has been supported by a recent survey~\cite{LearningFromRandomizedSimulators}, we encourage future works to extend the comparison to other ADR methods and settings. For instance, recent approaches have studied the underlying problem from a more general perspective, allowing to infer free-form source domain distributions given a starting data collection policy~\cite{BayesSim,neuralmuratore22}.

Finally, other works have approached the task in an unsupervised problem setting, i.e. \textit{without} making use of target domain data: \cite{ADR} drives policy training on progressively harder dynamics parameters to improve generalization, \cite{SPOTA} estimates the transfer performance with reference simulated environments, \cite{ADRrubikscube} encourages wider source domain distributions as long as the training performance is sufficiently high.

%% file: sections/03_methodology.tex
This section introduces the formal problem formulation, a detailed overview over online vs offline ADR methods, and our experimental setup.

\subsection{Problem setup}

\subsubsection{Problem formulation}

\input{sections/03a_problem_formulation_v1.tex}


\vspace{-12pt}
\subsubsection{Online ADR}
Online ADR methods, such as SimOpt~\cite{simopt2019} and BayRn~\cite{bayrn2021}, rely on iterative data collection from the target domain at optimization time (see Fig.~\ref{fig:on-vs-off_scheme}). They generally involve a bi-level optimization problem where RL policy training with DR in Eq.~\ref{eq:rl_goal} is interleaved by policy rollouts in the real setup.
At the $i$-th iteration, 
target transitions $\mathcal D_i$ are collected with the currently learned policy $\pi_{\phi_{i}}^{*}$ and used to adapt the source domain distribution $p_{\phi_{i+1}}(\xi)$, repeating the process until convergence.

While policies are still learned exclusively in simulation, these methods assume the real setup to be available for rolling out intermediate policies. Note how this assumption may be particularly expensive in robotic tasks, as careful expert supervision is needed when dealing with real hardware---e.g. for resetting the episodes, running safety checks, or interpreting and anticipating undesired behavior. Furthermore, intermediate policies may be the main point of failure themselves, as things can hardly be recovered if a policy happens to learn unwanted behavior in the middle of the process---e.g. it is unsafe to be run on the real setup, or it is not able to collect informative data, such as in the case of a robotic arm which moves through air without actually interacting with the environment.
Finally, these approaches may not be fully parallelized and run in a cluster in an end-to-end fashion, as previously noted by a related work~\cite{bayrn2021}.
\vspace{-12pt}
\subsubsection{Offline ADR}

Offline ADR methods make use of previously collected target domain data $\mathcal{D}$ to infer an optimal source domain distribution $p_{\phi^*}(\xi)$, and later train a single policy $\pi_{\phi^*}^{*}$ that can be directly transferred to the real world. An illustration of the pipeline of such methods is depicted in Fig.~\ref{fig:on-vs-off_scheme}, on the right.

In contrast to online methods, these make no assumptions on how real data has been collected. For this reason, offline ADR methods can be fed with human demonstrations, or data collected by previously trained policies for similar tasks. As such, previous studies in this line of work~\cite{droid2021,dropo2022} claim to adopt a safer and less restrictive pipeline. Note, indeed, how offline methods can potentially be run in online fashion---by collecting novel data with the newly learned policy and repeating the process---while the opposite is not always possible. Moreover, this approach allows for end-to-end parallelization---without expert personnel in the middle---and easier benchmarking of different algorithms by keeping a fixed shared dataset.
Despite the promising assumptions, these methods are by definition limited to the information provided by the previously collected trajectories, and may fail to readily adapt to the real domain.

\subsection{Benchmark methods}
\label{subsec:benchmark_methods}


\subsubsection{Uniform Domain Randomization}
We implement static domain randomization as a baseline, also known as Uniform Domain Randomization (UDR). Instead of manually tuning uniform distributions---which would be hard to fairly and systematically benchmark---we followed the same approach proposed by previous works~\cite{bayrn2021,dropo2022}: we average the results of 10 policies trained on randomly sampled uniform bounds, within a predefined search space over the dynamics parameters. This would essentially reflect the performances of a random search, allowing up to 10 evaluations.
\vspace{-14pt}
\subsubsection{Bayesian Domain Randomization (BayRn)}
Muratore et al. introduced BayRn~\cite{bayrn2021} as an affordable extension to UDR for adapting source domain distributions. BayRn optimizes the parameters $\phi$ through Bayesian Optimization (BO), using the final target domain return as optimization metric. By definition, BayRn falls under the area of online ADR methods.
We parameterize source distributions as uniform when implementing BayRn, and we initialize the Gaussian Process with the results of the 10 UDR policies, as suggested by the authors.
This makes BayRn slightly less comparable to the other ADR methods tested, which do not receive initial information. However, as we only allow a maximum of 5 real-world iterations in our benchmark protocol, we believe this step to be crucial for obtaining meaningful results through BO.
Note that BayRn does not explicitly use sensor measurements from the target domain, but critically requires reward computation on real hardware.
\vspace{-14pt}
\subsubsection{SimOpt}
SimOpt~\cite{simopt2019} uses a discrepancy metric between real and simulated trajectories to progressively optimize $p_{\phi}(\xi)$, with a trust region in the dynamics parameter space to stabilize the optimization process. Data is iteratively collected on the real setup with the currently trained policy, and later replayed in simulation with closed-loop actions from the same policy.
%
We implement SimOpt using the Relative Entropy Policy Search (REPS) implementation by~\cite{repsfinn2016}.
In order to maximize data efficiency, we only collect a single target trajectory per policy and perform multiple REPS updates at each iteration. A hyperparameter search is carried out over the number of intermediate REPS updates and number of sampled parameters per update, and we end up using 5 and 1000, respectively.

Finally, we implement a variant of SimOpt to investigate its performances when only a single real-world iteration is allowed, similarly to offline methods. Throughout the experiments, we refer to this variant as \textit{SimOpt-1}, which is implemented by executing the total number of REPS updates (25) during the first and only iteration.
\vspace{-14pt}
\subsubsection{DROID}
DROID~\cite{droid2021} proposes to use expert human demonstrations for adapting source domain distributions $p_{\phi}(\xi)$. Demonstrations are replayed in simulation by executing the same actions as in the real setup, allowing to regress dynamics parameters in simulation through CMA-ES~\cite{cmaes}.
As we only show experiments on torque-control tasks, we adapt the original algorithm to minimize the L2-norm between observations rather than joint torques. Moreover, we don't assume to provide optimal offline data for the underlying task, therefore we don't consider the penalty factor in the original objective function.
\vspace{-14pt}
\subsubsection{DROPO}
DROPO~\cite{dropo2022} proposes to optimize $p_{\phi}(\xi)$ with a maximum-likelihood approach, based on offline-collected data. Under this formulation, CMA-ES is used to regress both means and variances of the source domain distribution, by replaying offline data in simulation similarly to DROID. Nevertheless, DROPO further assumes that sensor measurements from the real setup give access to the full state configuration, in order to reset the simulated environment arbitrarily along the trajectory.
We test DROPO using the open-source implementation~\cite{dropo2022}, and tune the hyperparameter $\epsilon$ as suggested by the authors. Moreover, we set the sample size to be 10 times larger the dimensionality of tested dynamics parameter spaces, to allow for meaningful likelihood inference.

\subsection{Benchmark Tasks}
\label{sec:benchmark-tasks}

\begin{figure*}[t!]
\begin{tabular}{cccc}
\includegraphics[width=0.24\textwidth]{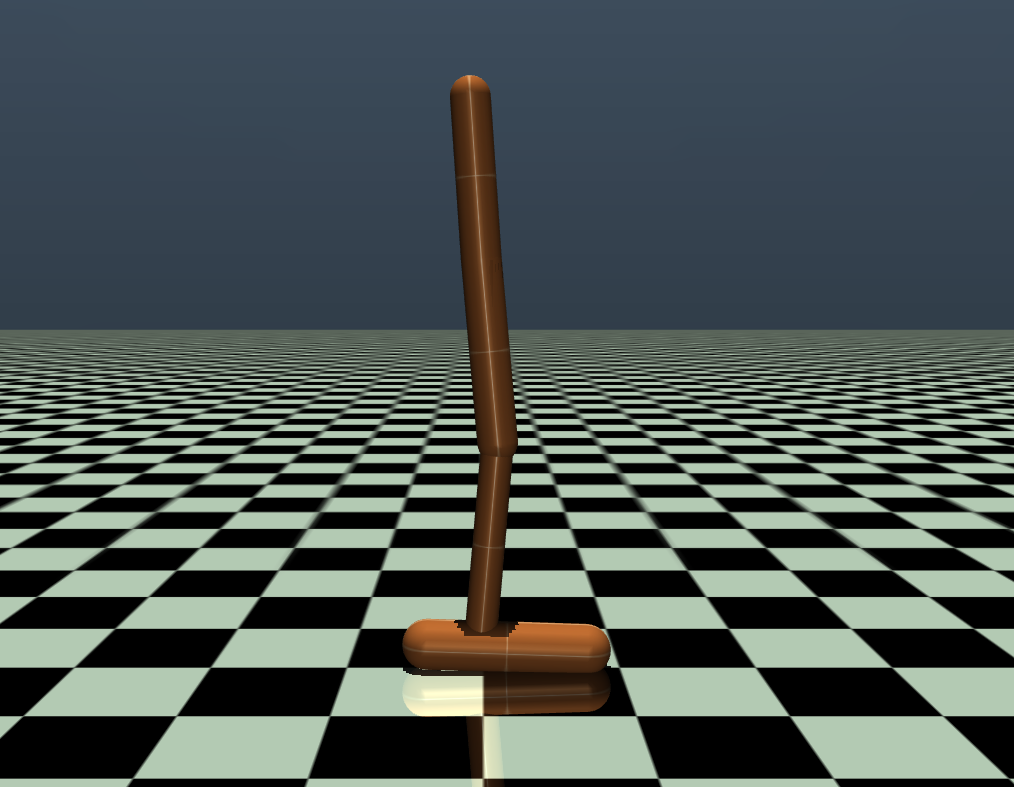} & \includegraphics[width=0.24\textwidth]{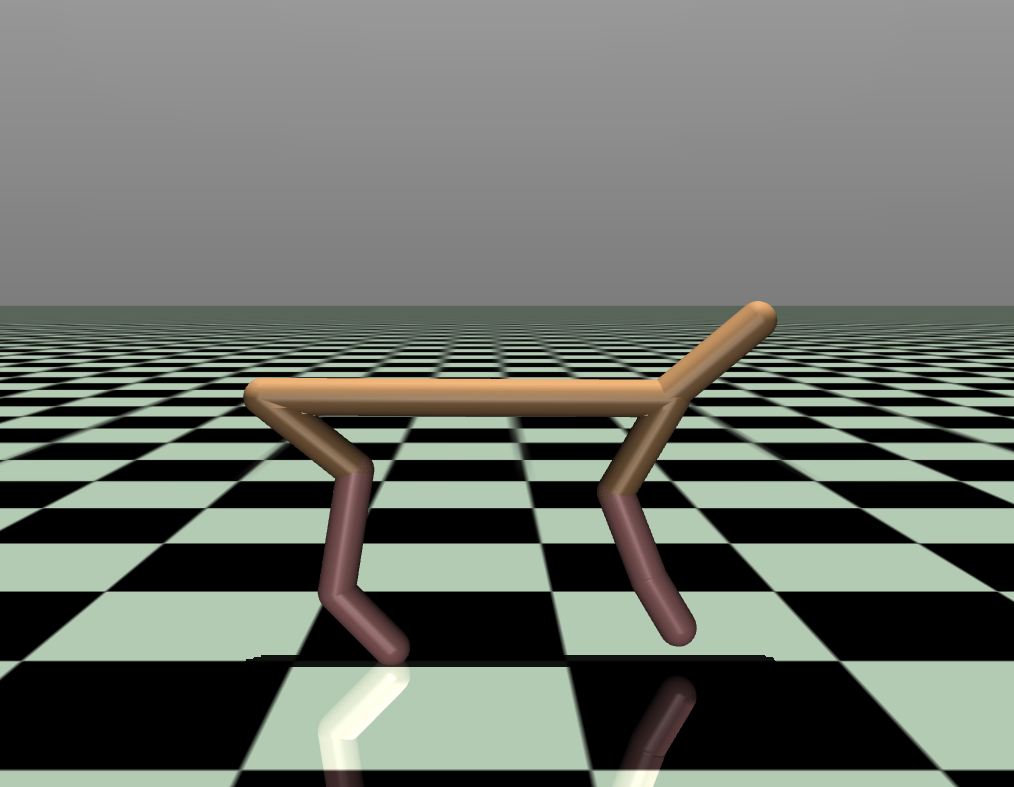} &
\includegraphics[width=0.24\textwidth]{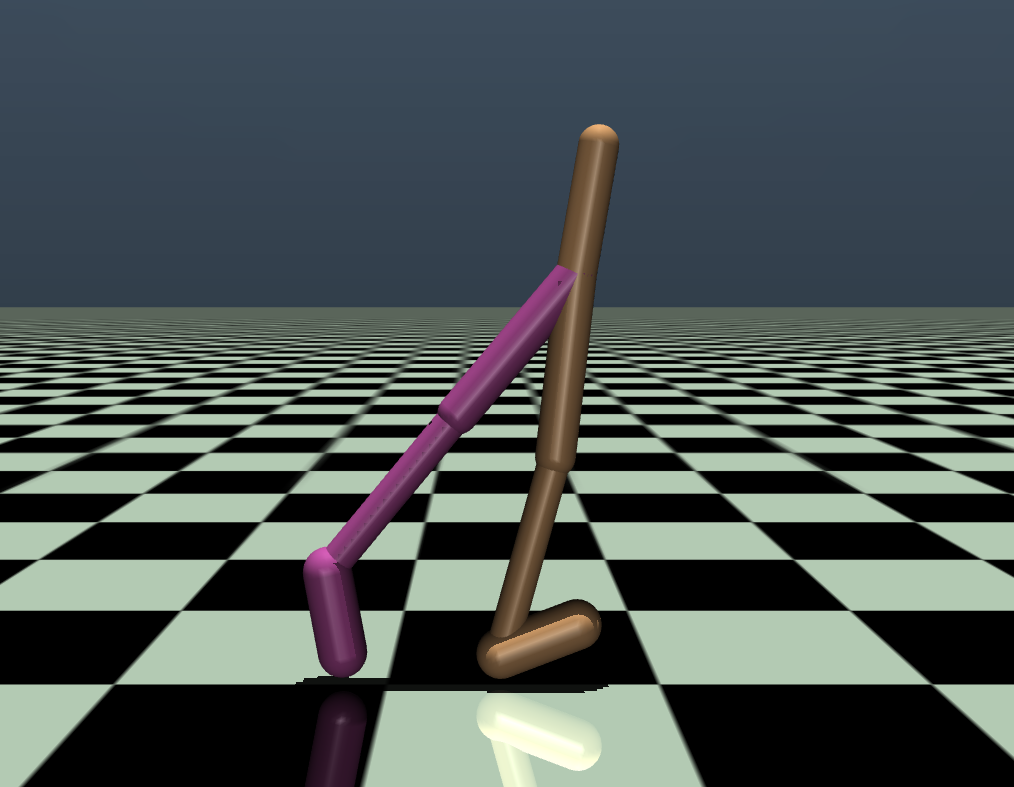} &
\includegraphics[width=0.24\textwidth]{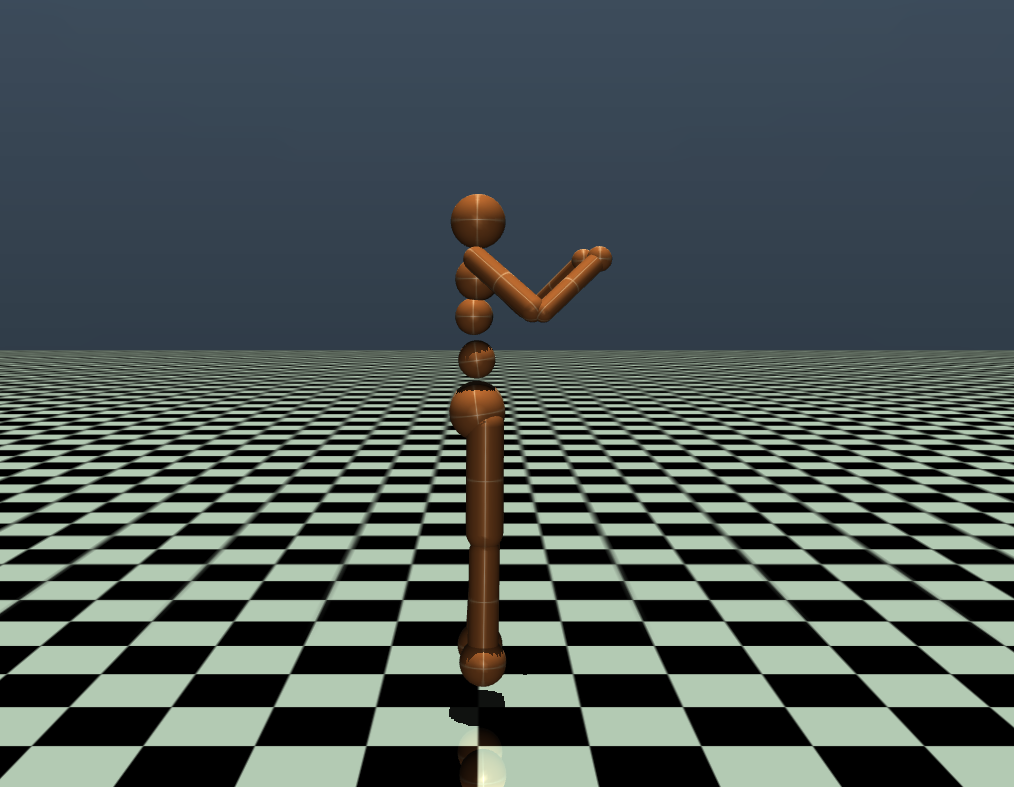} \\
(a) & (b) & (c) & (d)
\end{tabular}
\vspace{-0.3cm}
\caption{Illustration of the four OpenAI gym tasks used in this benchmark: Hopper (a), HalfCheetah (b), Walker2D (c), and Humanoid (d).}
\vspace{-0.3cm}
\label{fig:gym-environments}
\end{figure*}

We test the aforementioned ADR methods in four OpenAI gym~\cite{openai} environments, depicted in Figure.~\ref{fig:gym-environments}. These robotic tasks are commonly used to benchmark sim-to-real transfer methods in the absence of real hardware~\cite{AusersGuideForCalibratingSimulators}. To further simulate the reality gap, we consider each environment in three different versions: \textit{vanilla}, \textit{noisy}, and \textit{unmodeled}.
The first case corresponds to the basic condition with no mismatch between source and target.
The other two allow us to compare performances under noisy conditions---by injecting noise during target domain data collection---and in presence of unmodeled phenomena---by misidentifying a subset of dynamics parameters. More specifically, we simulate unmodeled phenomena as in~\cite{epopt,dropo2022}: a selected number of parameters is taken out of the optimization problem and underestimated by 20\% w.r.t. their original ground truth values. Note that the environments are designed by increasing difficulty, by progressively expanding the state space, dynamics parameter space, and number of unmodeled parameters. As the Humanoid environment contains a richer state space which includes masses and constraint forces, we only feed inference methods with the main bodies kinematics properties---i.e. the first 45 dimensions---while leaving full information to policies at training time. An overall description of the benchmark tasks is reported in Table~\ref{tab:benchmark_tasks}.

\input{tables/benchmark_tasks.tex}

\subsection{Benchmark Protocol}

As ADR methods share the same final goal, we propose a systematic framework to compare online vs. offline approaches by means of target domain policy performance and data efficiency. However, given the different assumptions and requirements highlighted in Sec.~\ref{subsec:benchmark_methods}, these metrics should by no means be considered as the sole axes of evaluation. Nevertheless, this work may help identifying which method is most suitable for a given problem setting, and how these algorithms perform when all requirements are fulfilled.

We benchmark online methods by allowing a maximum of 5 iterations. At each iteration, we limit data collection to a single trajectory clipped to a length of 200 transitions. This way, we keep the size of target datasets limited to a maximum of 1000 transitions, amounting to about 10 seconds-worth of locomotion data.
We feed offline methods with the cumulative trajectories collected by online methods up until the current iteration. In particular, we use data from SimOpt, as, similarly to DROID and DROPO, it makes use of sensory measurements from the target domain. By doing so, we ensure that SimOpt, DROID and DROPO have access to the same type and amount of information at all times.

Dynamics parameters are normalized to the interval $[0, 4]$, according to a predefined bounded search space. All methods start from a conservative prior source domain distribution centered in $2$ with identity covariance matrix. An exception to this is BayRn, which deals with uniform distributions and gets initialized with the results of the random search by UDR. In the case of the SimOpt single-iteration variant (SimOpt-1), we use the initial distribution to train a data collection policy, which is later used to collect the same amount of data used by offline methods, respectively for each iteration. We report the performances at the zero-th iteration as the target return of a policy trained on the initial conservative guess, marked as the starting point for all methods.

For policy training with Reinforcement Learning, we use the Soft Actor-Critic (SAC) implementation by \textit{stable-baselines3}~\cite{stable-baselines3}. We parametrized both actor and critic policies as 3-layer MLP neural networks, with 128 neurons in the hidden layers. We train each policy on 12 parallel environments, for a maximum of 5M timesteps overall. All SAC hyperparameters are kept at default values, except for the learning rate which is tuned separately for each environment by training on ground truth dynamics.
To ensure fair comparison among different policies, we stop the training process when a task-specific reward threshold is reached.

Finally, we report returns as the average performance over 3 random seeds---by repeating both the inference phase and policy training---normalized by the training reward threshold for the sake of clarity.

%% file: sections/03a_problem_formulation_v1.tex
Consider the source domain environment to be modeled as a discrete-time dynamical system, defined by a continuous state space $\mathcal{S} \in \mathbb{R}^{n_s}$, a continuous action space $\mathcal{A} \in \mathbb{R}^{n_a}$, and an initial state distribution $\mu(s_0)$.
In simulation, the environment can be further parameterized by its dynamics parameters $\xi \in \mathbb R^{n_\xi}$, e.g. masses, friction coefficients, robot link sizes.
The system dynamics are therefore described by the transition probability density function $\mathcal{P}_{\xi}(s_{t+1} | s_{t}, a_{t})$, given the current state $s_{t}$ and action taken $a_{t}$, at time step $t$. At each step, the agent is given a scalar reward feedback $r_t$ according to the function $R(s_t, a_t, s_{t+1})$, assumed to be deterministic for the sake of simplicity. Overall, the source environment forms a Markov Decision Process (MDP) described by the tuple $\mathcal M_{\xi} = \left<\mathcal S, \mathcal A, \mathcal P_{\xi}, \mu, R\right>$.
We assume $\xi$ to be random variables that obey a parametric distribution $p_{\phi}(\xi)$, parameterized by $\phi$---e.g. mean and variance for Gaussian distributions.

Under this formulation, the goal of an RL agent is to maximize the expected cumulative rewards (i.e. the return), by acting with a stochastic policy $\pi(a_t | s_t)$. In particular for the domain randomization setting, the goal is to train a robust policy over the source domain distribution $p_{\phi}(\xi)$,
\begin{equation}
\label{eq:rl_goal}
\pi_{\phi}^{*} =\underset{\pi }{\argmax } \ \mathbb{E}_{\xi \sim p_{\phi } (\xi )}\left[ E_{\pi, \mathcal{P}_{\xi } ,\mu _{0}}\left[ \ \sum _{t=0}^{T} \gamma ^{t} r_{t}\right]\right]
\end{equation}
with discount factor $\gamma \in (0,1]$. In practice, DR can be easily implemented into existing RL algorithms by sampling new dynamics parameters $\xi \sim p_{\phi}(\xi)$ at the start of each training episode, resulting in the outer expectation in Eq.~\ref{eq:rl_goal}.
ADR methods further include an inference phase in the overall pipeline to estimate the parameters $\phi$ of the source domain distribution $p_{\phi}(\xi)$. In this work, we consider Supervised ADR methods, i.e. approaches that use a collection of state-transitions $\mathcal{D} = \{(s_0, a_0, s_1, r_1), \dots , (s_{T-1}, a_{T-1}, s_T, r_T) \}$ from the target domain environment $\mathcal{M}_{real}$. In particular, we assume that the target MDP $\mathcal{M}_{real}$ shares the same state space, action space and reward function of $\mathcal M_{\xi}$, but generally differs by dynamics.

%% file: tables/benchmark_tasks.tex
\begin{table}[]
\begin{tabular}{l|cccccl}
\textbf{Environment} & \multicolumn{1}{l}{\textbf{\ \ $\mathcal S$ \ \ }} & \multicolumn{1}{l}{\textbf{\ $\mathcal A$  \ \  }} & \multicolumn{1}{l}{\textbf{\ $\xi$ \ }} & \multicolumn{1}{l}{\textbf{\ Unmod. $\xi$ \ }} & \multicolumn{1}{l}{\textbf{\ Noise \ }} & \textbf{Parameters}               \\ \hline
Hopper               & 11                                             & 3                                              & 4                                      & 1                                             & $10^{-4}$                          & Link masses                       \\ \hline
HalfCheetah          & 17                                             & 6                                              & 8                                      & 3                                             & $10^{-4}$                          & Link masses, friction             \\ \hline
Walker2d             & 17                                             & 6                                              & 13                                     & 4                                             & $10^{-3}$                          & Link masses and lengths, friction \\ \hline
Humanoid             & 376                                            & 17                                             & 30                                     & 7                                             & $10^{-3}$                          & Link masses, joint damping        \\ \hline
\end{tabular}
\vspace{6pt}
\caption{List of OpenAI gym~\cite{openai} environments used in this benchmark. The dimensionality of the state space, action space, dynamics parameter space, and unmodeled parameter space is reported, together with the noise level.}
\label{tab:benchmark_tasks}
    \vspace{-1.0cm}
\end{table}


%% file: sections/04_results.tex
\subsection{Vanilla Parameter Estimation}

\begin{figure*}[t!]
\includegraphics[width=\textwidth]{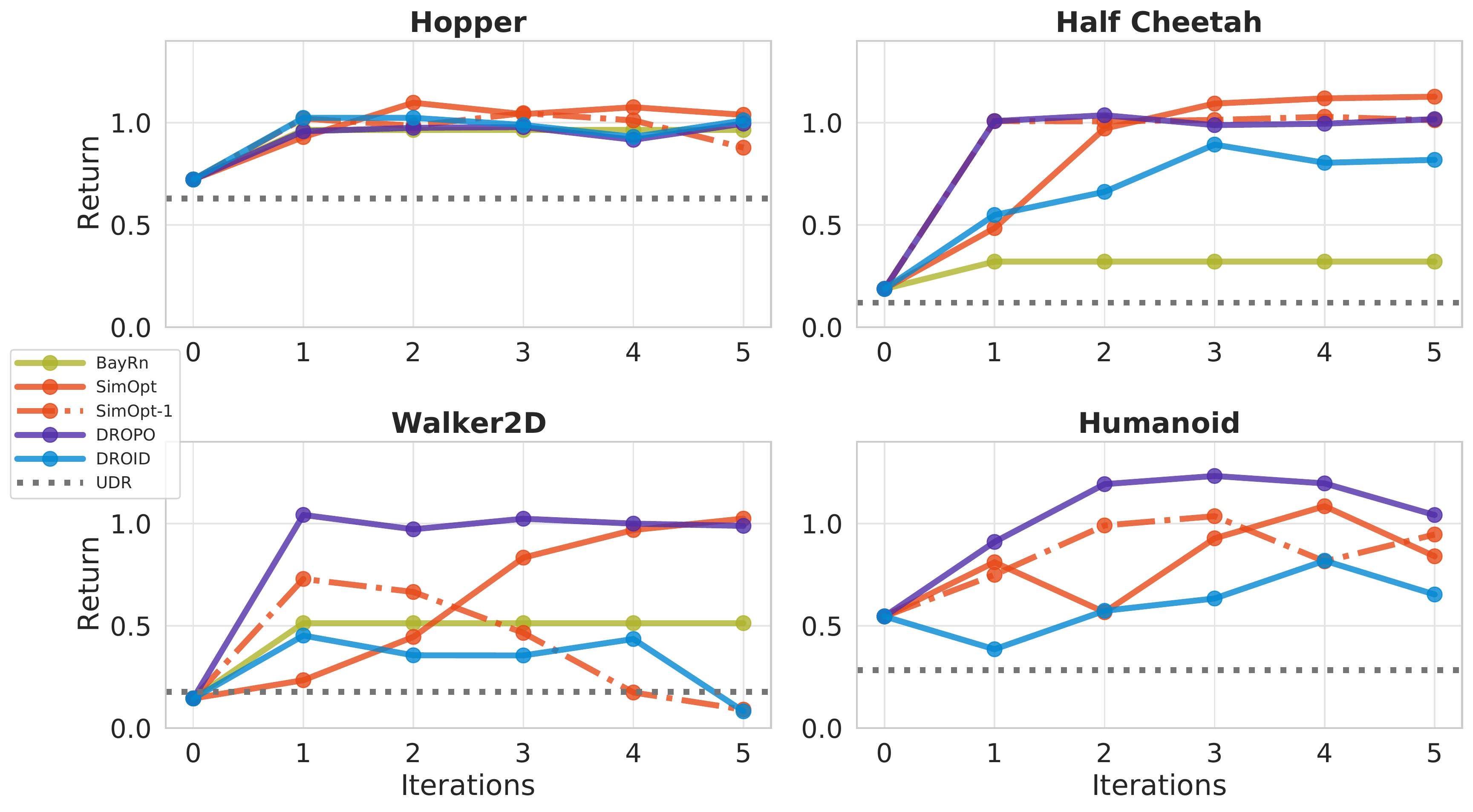}
    \vspace{-0.6cm}
    \caption{Policy returns in the target domain (normalized by training reward threshold) in the vanilla parameter estimation sim-to-sim setting.}
    \label{fig:vanilla-results}
    \vspace{-0.75cm}
\end{figure*}

As a starting point, we benchmark the outlined methods in a direct parameter estimation setting. Under this formulation, the two domains are solely mismatched by discrepancies in dynamics parameter values. ADR methods thus attempt to estimate source domain distributions that resemble the ground truth parameters, by collecting data from the target environment. While previous works mostly investigated the inference results under this setting~\cite{AusersGuideForCalibratingSimulators}, we directly focus on the policy performance in the ground truth environment, with respect to the amount of data seen by each algorithm.

The results are depicted in Fig.~\ref{fig:vanilla-results}, in terms of returns normalized by the training reward threshold. We observe that all methods are able to successfully transfer in the low-dimensional Hopper task, while some discrepancy is noted for more complex environments. In particular, we found that open-loop replaying of target trajectories by DROID generally leads to less stable results, with respect to SimOpt's approach of collecting sim trajectories by rolling out the currently converged policy. This effect is particularly visible in the Walker2D task where multiple trajectories lead to lower performances for DROID, e.g. if the Walker-agent happens to fall down while executing the target actions, the trajectory discrepancy metric may get uninformative. SimOpt authors have stated similar observations when experimenting with open-loop replaying in their experiments (See Appendix A. in~\cite{simopt2019}). We suspect that the trust region imposed by SimOpt may further prevent diverging trajectories, as the next-iteration parameter search likely falls closer to the current policy training dynamics w.r.t. CMA-ES unconstrained search. Moreover, we noticed that, while executing actions in the same way as DROID, DROPO does not suffer from this. To this end, we found that intermediately resetting the simulator state is crucial to prevent sim trajectories to diverge from real ones, and provide informative distance measures. 

Overall, we observe that SimOpt is able to progressively get better results and solve the task in all cases within 5 iterations, i.e. with 5 or less trajectories collected from the target domain. Interestingly enough, the quality of the policy also matters for SimOpt, as SimOpt-1 fails in the Walker2D task when collecting all trajectories with the poorly performant initial policy.
On the other hand, BayRn fell short on higher-dimensional tasks when only 5 iterations are allowed, as Gaussian Processes are notoriously problematic to scale to high parameter dimensions ($\geq8$ in our case). Due to this, we do not report results for Bayesian Optimization on the Humanoid environment.

Finally, we found that DROPO is able to solve all tasks and match the long-term results of SimOpt in this setting, even with a single target trajectory.

\subsection{Observation noise}
We test the transfer performance of all benchmark methods when the target domain is assumed to provide noisy measurements, as in the case of real hardware. For this purpose, we feed each algorithm with target observations containing injected noise according to the predefined values in Table~\ref{tab:benchmark_tasks}.

\begin{figure*}[t!]
\includegraphics[width=\textwidth]{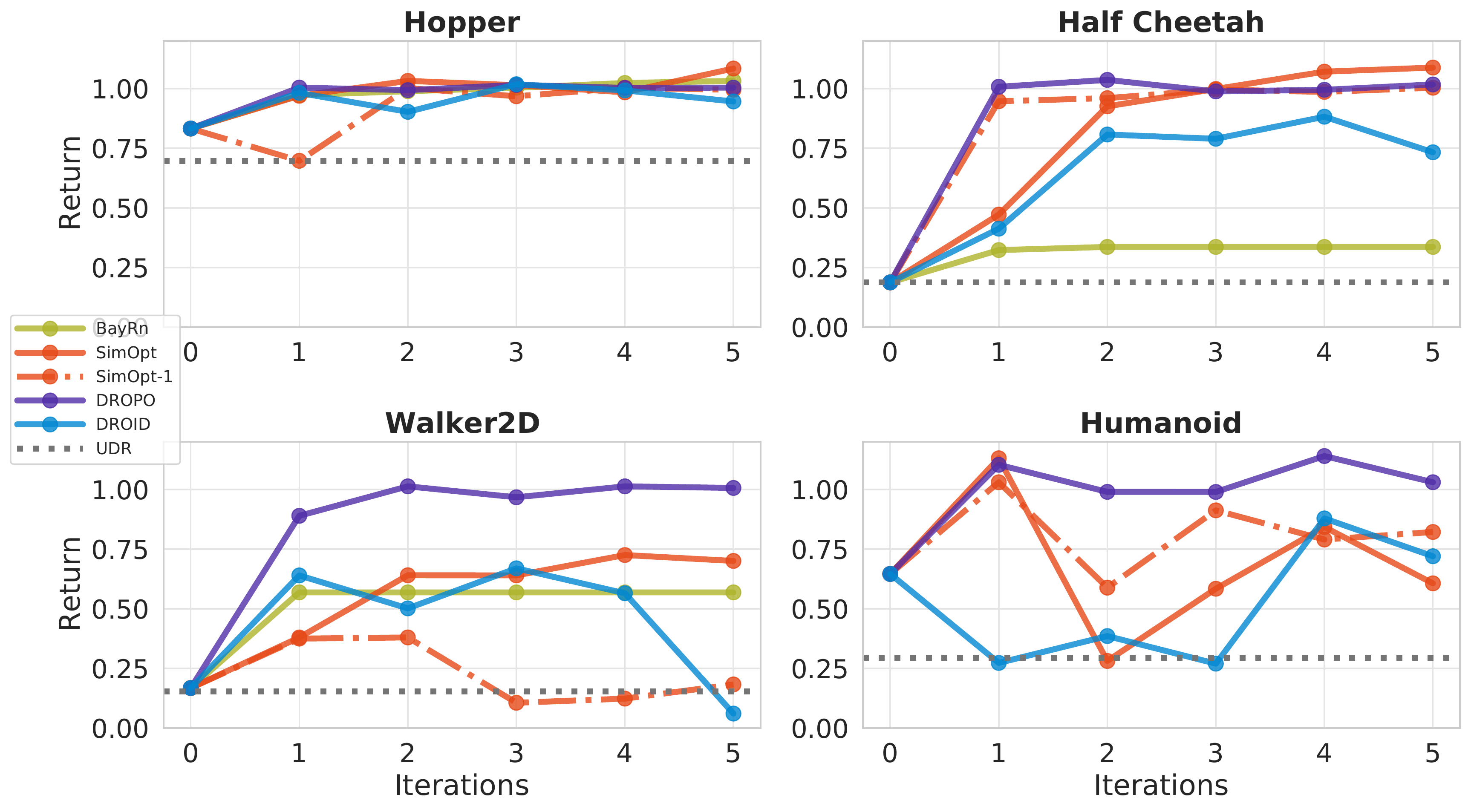}
    \vspace{-0.6cm}
    \caption{Normalized returns in the target domain under noisy observations.}
    \label{fig:noisy-results}
    \vspace{-0.5cm}
\end{figure*}

We report the results for this setting in Fig.~\ref{fig:noisy-results}. In general, we draw similar conclusion as in the direct parameter estimation setting: (i) all methods succeed on the Hopper task, (ii) DROID and SimOpt-1 may fail unexpectedly when sim trajectories diverge during replay (see Walker2D in Fig.~\ref{fig:noisy-results}), (iii) BayRn fails to transfer on problems with higher-dimensional dynamics parameter spaces. Moreover, in this setting we experienced generally lower SimOpt performances, especially for harder tasks: we found that SimOpt's intermediate policies may at times fail to transfer in noisy conditions and hinder the overall iterative process (see SimOpt on Humanoid in Fig.~\ref{fig:noisy-results} after the first iteration). This is in line with the statements by~\cite{bayrn2021} during their experimental evaluation.

We finally observe that DROPO is still able to transfer across all tasks given a single trajectory of 200 transitions, showing slight improvements as more noisy data becomes available.

\subsection{Unmodeled phenomena} 
Finally, we report each method's performance in presence of unmodeled phenomena
(see Sec.~\ref{sec:benchmark-tasks}). This approach would reflect a more realistic transfer scenario where not all physical parameters are considered during the inference phase and potentially mismatched from true real-world values.

\begin{figure*}[t!]
\includegraphics[width=\textwidth]{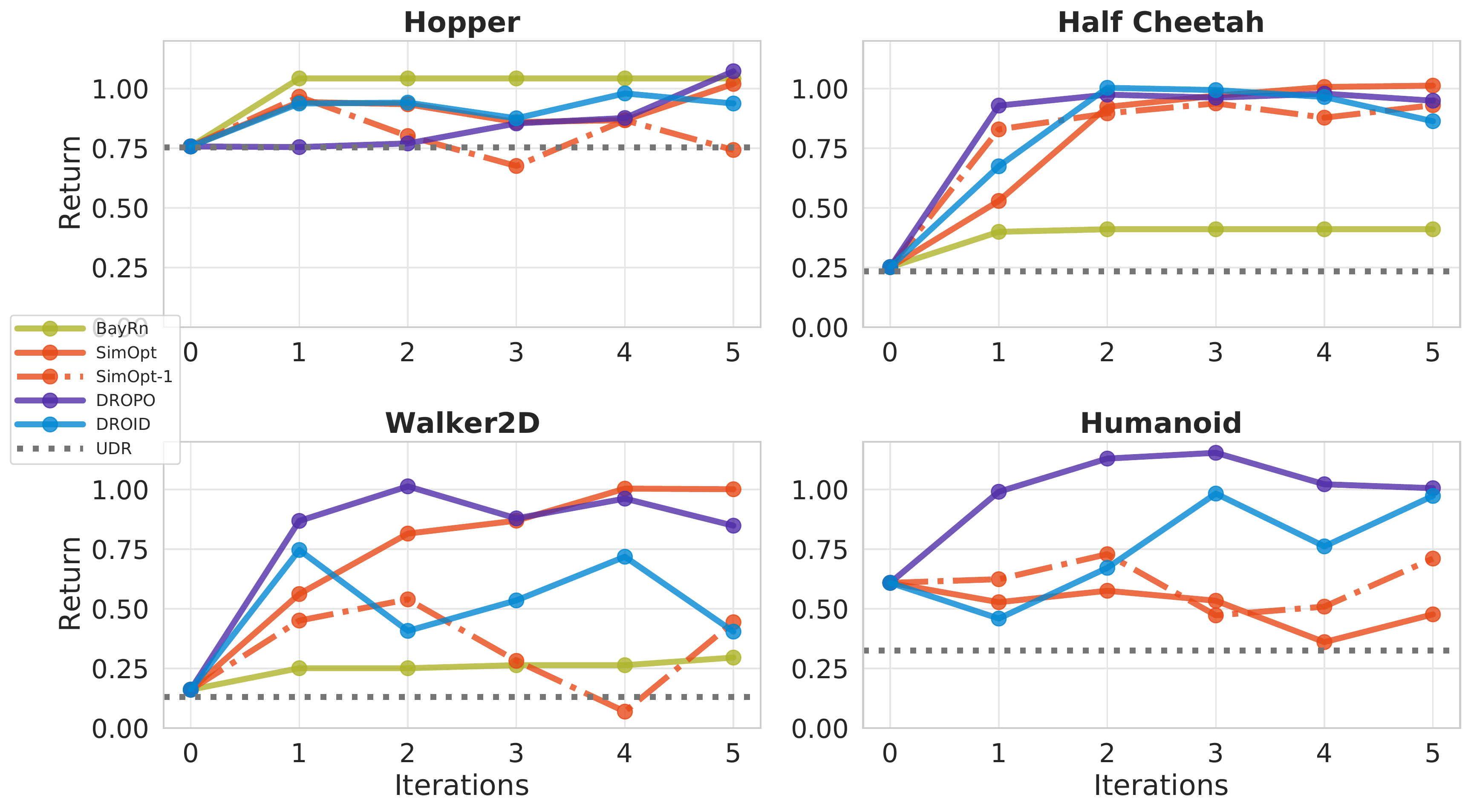}
    \vspace{-0.6cm}
    \caption{Normalized returns in the target domain in presence of unmodeled phenomena.}
    \label{fig:unmodeled-results}
        \vspace{-0.5cm}
\end{figure*}

The results are depicted in Fig.~\ref{fig:unmodeled-results}. In comparison to the vanilla parameter estimation setting, we observe that offline ADR methods behaved similarly, while SimOpt and its single-iteration variant struggled to solve the task in the Humanoid environment. We believe this to be caused by intermediate ill-performing policies which prevented a successful parameter estimation and hindered the overall process. Besides this occurrence, SimOpt demonstrated stable and successful transfers within 5 trajectories in the remaining tasks.
Surprisingly, BayRn was shown to obtain superior performance in the Hopper task than other methods, likely by avoiding explicit parameter estimation and directly optimizing for target domain returns. Overall, we found DROPO to be the best performing method under this setting, despite requiring significantly more resources for hyperparameter tuning: DROPO inherently requires multiple runs to decide on the best regularization value $\epsilon$, which was found to significantly impact on final performances. Nevertheless, we stuck to the original suggested procedure for tuning this parameter~\cite{dropo2022}. In addition, we experienced noticeably higher inference times by DROPO, as likelihood estimation requires a considerable amount of Monte-Carlo sampling. On the other hand, DROID can be run with minimal hyperparameter tuning and faster objective function evaluations, but still generally suffered from divergent trajectories during parameter search on the harder tasks. 
Furthermore, we observed that DROID consistently converged to nearly zero-variance distributions, as a result of the CMA-ES covariance matrix decaying to infinitesimal values after finding a local optimum.
This behavior limits the potential of domain randomization, effectively reducing it to system identification.

\subsection{Off-policy data collection}
\label{sec:offpolicy_data_collection}

\begin{figure*}[h]
\centering
\includegraphics[width=\textwidth]{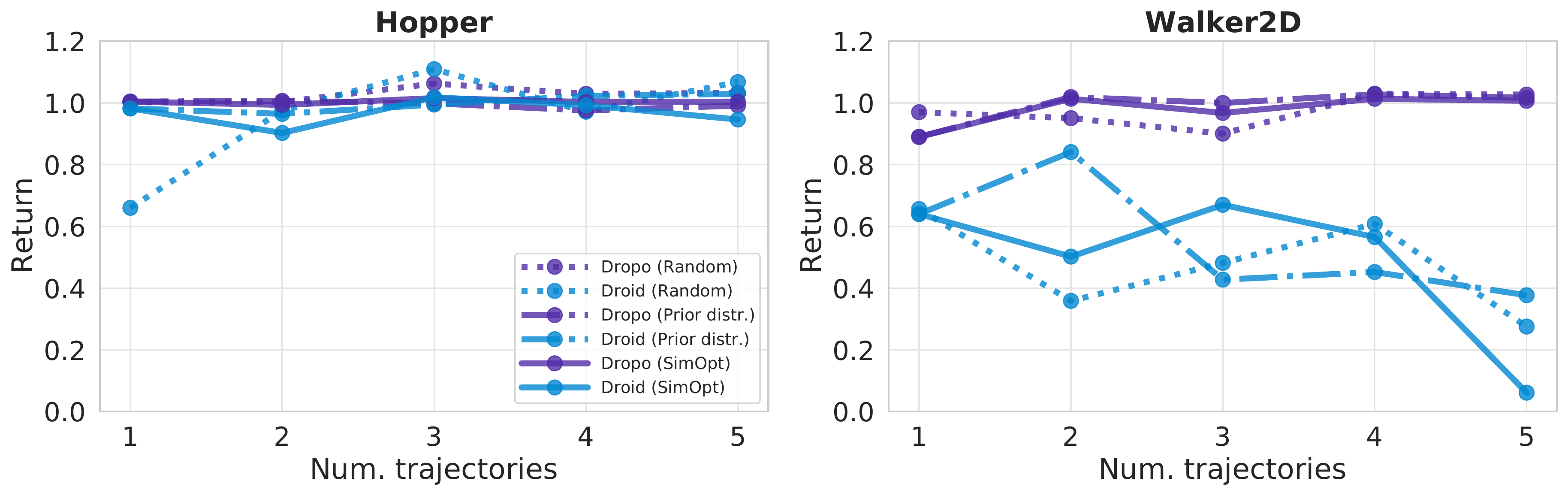}
    \vspace{-0.6cm}
    \caption{Target domain performance with different data collection strategies under noisy conditions.}
    \label{fig:datacoll-results}
    \vspace{-0.5cm}
\end{figure*}

We designed the main experimental evaluation by feeding offline methods with the cumulative data collected by SimOpt, in order to compare the methods with the same target domain trajectories. As this would hardly be the case in a real application, in this section we investigate how different data collection strategies impact on offline ADR performances. In particular, we test how target returns with data from SimOpt compare to (i) randomly collected data, and (ii) data collected with a policy trained on the prior source distribution $\mathcal N(\mathbf{2}, \mathds{1} )$---as in the first iteration of the main results. We test the performances in the Hopper and Walker2D tasks, under noisy conditions, and report the results in Fig.~\ref{fig:datacoll-results}.

We observe that different strategies did not significantly affect the performances in the Hopper environment when sufficient transitions are provided. However, we noticed that DROID is significantly more sensitive to the quality of collected data in the Walker2D task, with respect to DROPO. As previously argued, we attribute this phenomenon to DROID's open-loop replaying of target actions in simulation, which may lead to uninformative distance measures---e.g. when the Walker2D-agent falls down, or when the sim trajectories diverge from the target one due to highly mismatched dynamics parameters. Nevertheless, we claim that tasks different from locomotion may pose further requirements to data collection procedures, e.g. robotic manipulation would at least require the robot to interact with the objects in the scene.

Finally note that, while we could not test this strategy in sim-to-sim settings, offline methods may still be fed with data collected by kinesthetic guidance---i.e. human demonstrations---as opposed to online methods.

%% file: sections/05_conclusions.tex
In this paper we provide an extensive experimental evaluation of four Supervised Adaptive Domain Randomization methods in terms of target policy performance and data efficiency, when tested on challenging sim-to-sim transfer scenarios. Additionally, we highlight the different assumptions and requirements of each benchmark method, while splitting them into online vs. offline settings w.r.t. to data collection procedures and optimization pipelines.
We found that online methods are able to solve most target tasks within 5 data collection iterations---i.e. 10 seconds-worth of data---with least hyperparameter tuning and inference time. Interestingly, we observed that their performance is affected by the quality of the currently learned policy, as they may sometimes fail unexpectedly due to intermediate ill-behaving policies, posing it as the main limitation of online methods. On the other hand, offline methods generally led to better jump-start performance with fewer target transitions available, even when compared to a single iteration of SimOpt. In particular, DROPO achieved the highest average return among the tested tasks, given stricter assumptions---i.e. full knowledge of real-world state configurations is assumed. In contrast, we observed that open-loop replaying of trajectories by DROID may lead to divergent sim trajectories and, in turn, less informative trajectory discrepancy measures.

This work provides the first empirical insights into the performances of current ADR algorithms with respect to the type and amount of data required. These considerations can help researchers in the field to continue on improving sim-to-real transfer methods with Domain Randomization, and make informed decisions when applying these in future works.

%% file: main.bbl
\begin{thebibliography}{10}
\providecommand{\url}[1]{{#1}}
\providecommand{\urlprefix}{URL }
\expandafter\ifx\csname urlstyle\endcsname\relax
  \providecommand{\doi}[1]{DOI~\discretionary{}{}{}#1}\else
  \providecommand{\doi}{DOI~\discretionary{}{}{}\begingroup
  \urlstyle{rm}\Url}\fi

\bibitem{pivoting}
Antonova, R., Cruciani, S., Smith, C., Kragic, D.: Reinforcement learning for
  pivoting task.
\newblock arXiv Preprint \textbf{1703.00472} (2017)

\bibitem{openai}
Brockman, G., Cheung, V., Pettersson, L., Schneider, J., Schulman, J., Tang,
  J., Zaremba, W.: Openai gym.
\newblock arXiv Preprint \textbf{1606.01540} (2016)

\bibitem{simopt2019}
Chebotar, Y., Handa, A., Makoviychuk, V., Macklin, M., Issac, J., Ratliff, N.,
  Fox, D.: Closing the sim-to-real loop: Adapting simulation randomization with
  real world experience.
\newblock In: ICRA (2019)

\bibitem{chen2022understanding}
Chen, X., Hu, J., Jin, C., Li, L., Wang, L.: Understanding domain randomization
  for sim-to-real transfer.
\newblock In: ICLR (2022)

\bibitem{tactilesensory}
Ding, Z., Tsai, Y., Lee, W.W., Huang, B.: Sim-to-real transfer for robotic
  manipulation with tactile sensory.
\newblock In: {IROS} (2021)

\bibitem{repsfinn2016}
Finn, C., Zhang, M., Fu, J., Tan, X., McCarthy, Z., Scharff, E., Levine, S.:
  Guided policy search code implementation (2016).
\newblock \urlprefix\url{http://rll.berkeley.edu/gps}.
\newblock Software available from rll.berkeley.edu/gps

\bibitem{cmaes}
Hansen, N.: The CMA Evolution Strategy: A Comparing Review (2006)

\bibitem{James2017}
James, S., Davison, A., Johns, E.: Transferring end-to-end visuomotor control
  from simulation to real world for a multi-stage task.
\newblock pp. 334--343. PMLR (2017)

\bibitem{kober2013reinforcement}
Kober, J., Bagnell, J.A., Peters, J.: Reinforcement learning in robotics: A
  survey.
\newblock The International Journal of Robotics Research \textbf{32}(11),
  1238--1274 (2013)

\bibitem{ADR}
Mehta, B., Diaz, M., Golemo, F., Pal, C.J., Paull, L.: Active domain
  randomization.
\newblock In: CoRL (2020)

\bibitem{AusersGuideForCalibratingSimulators}
Mehta, B., Handa, A., Fox, D., Ramos, F.: A user‘s guide to calibrating
  robotics simulators.
\newblock In: CoRL (2020)

\bibitem{bayrn2021}
Muratore, F., Eilers, C., Gienger, M., Peters, J.: Data-efficient domain
  randomization with bayesian optimization.
\newblock {IEEE} ICRA - Robotics Autom. Lett. \textbf{6}(2), 911--918 (2021)

\bibitem{SPOTA}
Muratore, F., Gienger, M., Peters, J.: Assessing transferability from
  simulation to reality for reinforcement learning.
\newblock IEEE TPAMI \textbf{43}(4), 1172--1183 (2021)

\bibitem{neuralmuratore22}
Muratore, F., Gruner, T., Wiese, F., Belousov, B., Gienger, M., Peters, J.:
  Neural posterior domain randomization.
\newblock In: A.~Faust, D.~Hsu, G.~Neumann (eds.) Proceedings of the 5th
  Conference on Robot Learning, \emph{Proceedings of Machine Learning
  Research}, vol. 164, pp. 1532--1542. PMLR (2022)

\bibitem{LearningFromRandomizedSimulators}
Muratore, F., Ramos, F., Turk, G., Yu, W., Gienger, M., Peters, J.: Robot
  learning from randomized simulations: {A} review.
\newblock Frontiers Robotics {AI} \textbf{9}, 799,893 (2022)

\bibitem{ADRrubikscube}
{OpenAI}, Akkaya, I., Andrychowicz, M., Chociej, M., Litwin, M., McGrew, B.,
  Petron, A., Paino, A., Plappert, M., Powell, G., Ribas, R., Schneider, J.,
  Tezak, N., Tworek, J., Welinder, P., Weng, L., Yuan, Q., Zaremba, W., Zhang,
  L.: Solving rubik's cube with a robot hand.
\newblock arXiv Preprint \textbf{1910.07113} (2019)

\bibitem{dynamics2018}
Peng, X.B., Andrychowicz, M., Zaremba, W., Abbeel, P.: Sim-to-real transfer of
  robotic control with dynamics randomization.
\newblock In: ICRA (2018)

\bibitem{stable-baselines3}
Raffin, A., Hill, A., Gleave, A., Kanervisto, A., Ernestus, M., Dormann, N.:
  Stable-baselines3: Reliable reinforcement learning implementations.
\newblock Journal of Machine Learning Research \textbf{22}(268), 1--8 (2021).
\newblock \urlprefix\url{http://jmlr.org/papers/v22/20-1364.html}

\bibitem{epopt}
Rajeswaran, A., Ghotra, S., Ravindran, B., Levine, S.: Epopt: Learning robust
  neural network policies using model ensembles.
\newblock In: {ICLR} (2017)

\bibitem{BayesSim}
Ramos, F., Possas, R.C., Fox, D.: Bayessim: adaptive domain randomization via
  probabilistic inference for robotics simulators.
\newblock In: {RSS} (2019)

\bibitem{DRvision2017rss}
Sadeghi, F., Levine, S.: {CAD2RL:} real single-image flight without a single
  real image.
\newblock In: RSS (2017)

\bibitem{locomotion}
Tan, J., Zhang, T., Coumans, E., Iscen, A., Bai, Y., Hafner, D., Bohez, S.,
  Vanhoucke, V.: Sim-to-real: Learning agile locomotion for quadruped robots.
\newblock In: RSS (2018)

\bibitem{dropo2022}
Tiboni, G., Arndt, K., Kyrki, V.: {DROPO:} sim-to-real transfer with offline
  domain randomization.
\newblock arXiv Preprint \textbf{2201.08434} (2022)

\bibitem{DRvision2017iros}
Tobin, J., Fong, R., Ray, A., Schneider, J., Zaremba, W., Abbeel, P.: Domain
  randomization for transferring deep neural networks from simulation to the
  real world.
\newblock In: {IROS} (2017)

\bibitem{droid2021}
Tsai, Y., Xu, H., Ding, Z., Zhang, C., Johns, E., Huang, B.: {DROID:}
  minimizing the reality gap using single-shot human demonstration.
\newblock {IEEE} Robotics Autom. Lett. \textbf{6}(2), 3168--3175 (2021)

\bibitem{Valassakis}
Valassakis, E., Di~Palo, N., Johns, E.: Coarse-to-fine for sim-to-real:
  Sub-millimetre precision across wide task spaces.
\newblock In: IROS (2021)

\bibitem{howtopickdrdistributions}
Vuong, Q., Vikram, S., Su, H., Gao, S., Christensen, H.: {H}ow to pick the
  domain randomization parameters for sim-to-real transfer of reinforcement
  learning policies?
\newblock In: ICRA (2019)

\bibitem{zhao2020sim}
Zhao, W., Queralta, J.P., Westerlund, T.: Sim-to-real transfer in deep
  reinforcement learning for robotics: a survey.
\newblock In: 2020 IEEE Symposium Series on Computational Intelligence (SSCI),
  pp. 737--744. IEEE (2020)

\end{thebibliography}
